\documentclass[journal]{IEEEtran}
\usepackage{ifpdf}

\usepackage{cite}

\ifCLASSINFOpdf
   \usepackage[pdftex]{graphicx}
\else
   \usepackage[dvips]{graphicx}
\fi

\usepackage{booktabs}
\usepackage{multirow}
\usepackage{multicol}

\usepackage[cmex10]{amsmath}
\usepackage{algorithm}

\usepackage{algorithmic}

\usepackage{array}

\usepackage[numbers]{natbib}

\ifCLASSOPTIONcompsoc
  \usepackage[caption=false,font=normalsize,labelfont=sf,textfont=sf]{subfig}
\else
  \usepackage[caption=false,font=footnotesize]{subfig}
\fi

\usepackage{hyperref}

\newtheorem{definition}{Definition}

\begin{document}
%
\title{AdaResNet: Enhancing Residual Networks with Dynamic Weight Adjustment for Improved Feature Integration}
%
%
%

\author{Hong~Su 
\IEEEcompsocitemizethanks{\IEEEcompsocthanksitem H. Su is with School of Computer Science, Chengdu University of Information Technology, Chengdu, China.\protect\\
E-mail: suguest@126.com
E-mail: xinhua@139.com}
\thanks{}}

\markboth{IEEE LaTeX Version,~Vol.~X, No.~X, X~X}%
{Shell \MakeLowercase{\textit{et al.}}: Bare Demo of IEEEtran.cls
for Journals}

\maketitle

\begin{abstract}
    In very deep neural networks, gradients can become extremely small during backpropagation, making it challenging to train the early layers. ResNet (Residual Network) addresses this issue by enabling gradients to flow directly through the network via skip connections, facilitating the training of much deeper networks. However, in these skip connections, the input (\( \textit{ipd} \)) is directly added to the transformed data (\( \textit{tfd} \)), treating \( \textit{ipd} \) and \( \textit{tfd} \) equally, without adapting to different scenarios. In this paper, we propose AdaResNet (Auto-Adapting Residual Network), which automatically adjusts the ratio between \( \textit{ipd} \) and \( \textit{tfd} \) based on the training data. We introduce a variable, \( \textit{weight}_{tfd}^{ipd} \), to represent this ratio. This variable is dynamically adjusted during backpropagation, allowing it to adapt to the training data rather than remaining fixed. Experimental results demonstrate that AdaResNet achieves a maximum accuracy improvement of over 50\% compared to traditional ResNet.
\end{abstract}


%
\IEEEpeerreviewmaketitle

\section{Introduction}
\IEEEPARstart{I}{n} recent years, deep learning has revolutionized numerous fields, ranging from computer vision and natural language processing to autonomous systems and beyond. Among the various architectures that have emerged, ResNet (Residual Network) has played a pivotal role in advancing the state of the art in these domains \cite{xu2023resnet} \cite{anand2024enhanced}. Its innovative design has enabled the training of extremely deep neural networks by addressing a critical challenge faced in traditional deep architectures: the vanishing gradient problem.

As neural networks become deeper, gradients can diminish significantly during the backpropagation process. This issue hampers the effective training of the early layers, causing the network to stagnate and preventing it from learning meaningful representations. ResNet tackles this problem by introducing skip connections \cite{weng2024tailor}, which allow gradients to bypass intermediate layers and flow directly through the network. This mechanism facilitates the training of much deeper networks, making it possible to achieve unprecedented levels of accuracy and performance on complex tasks.

\begin{figure}[htb]
    \includegraphics[width = 4 in]{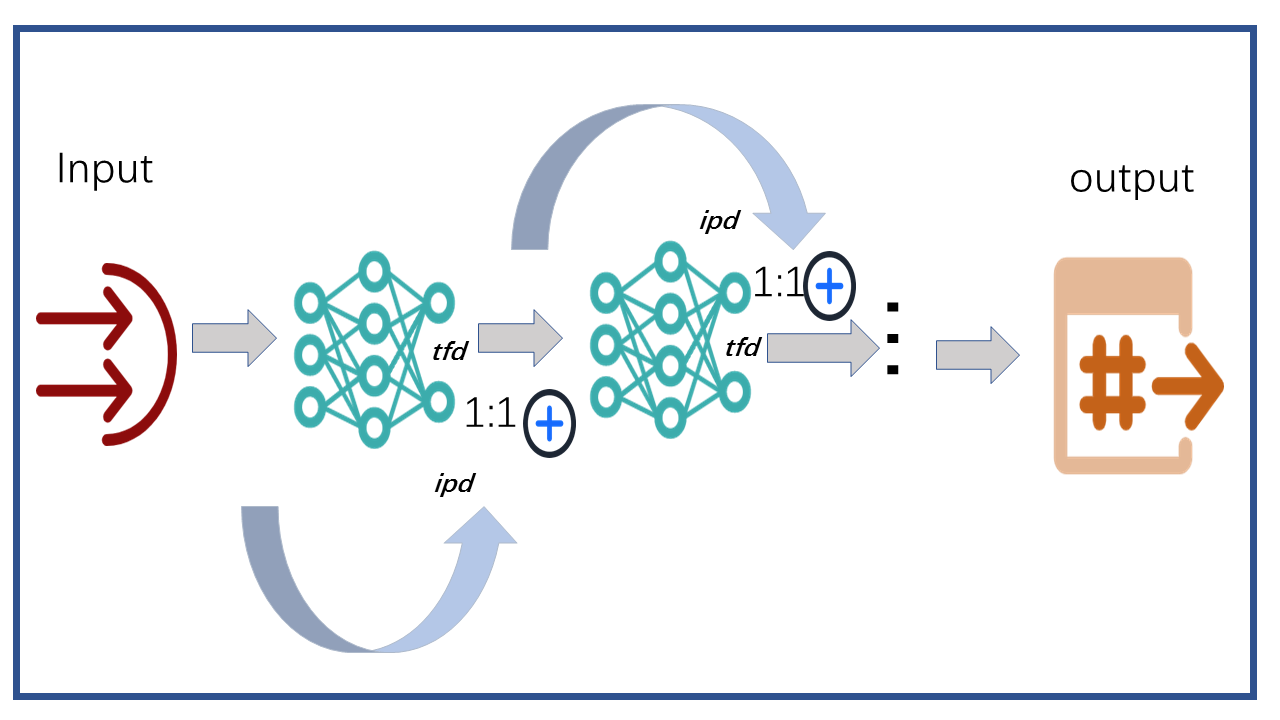}
    \caption{ResNet to add the input and intermediately processed directly to increase gradients for deep neural network}
    \label{whyResNET}
\end{figure}

Despite the success of ResNet, the standard implementation of skip connections involves directly adding the input ($ipd$) to the transformed data ($tfd$), i.e., they are combined in a fixed ratio of 1:1, as illustrated in Figure \ref{whyResNET}. This approach inherently assumes that $ipd$ and $tfd$ contribute equally to the network's output, which may not be optimal across all recognition scenarios. By treating $ipd$ and $tfd$ as identical in their contribution, the traditional ResNet architecture does not account for the varying importance of $ipd$ and $tfd$ across different layers or diverse training data distributions.

In this paper, we propose a novel architecture, AdaResNet (Auto-Adapting Residual Network), which enhances the flexibility of ResNet by automatically adapting the contribution of $ipd$ and $tfd$ during training. Specifically, we introduce a learnable parameter, denoted as $weight_{tfd}^{ipd}$, which dynamically adjusts the ratio between $ipd$ and $tfd$ based on the training data. Unlike traditional ResNet, where the combination of $ipd$ and $tfd$ remains fixed, AdaResNet allows this ratio to be tuned throughout the training process, thereby improving the network's ability to generalize across diverse data distributions.

The contributions of this paper are threefold:
(1) Introduction of AdaResNet: We present AdaResNet, a novel extension of the ResNet architecture that incorporates an adaptive mechanism to balance the contributions of skipped input ($ipd$) and processed data ($tfd$). This approach overcomes the limitations of the fixed 1:1 ratio combination used in traditional ResNet, allowing for more flexible and effective integration of $ipd$ and $tfd$.
(2) Learnable parameter \(weight_{tfd}^{ipd}\): We propose a new learnable parameter, \(weight_{tfd}^{ipd}\), which is automatically optimized during training. This parameter enables the network to dynamically adjust the balance between $ipd$ and $tfd$ in response to varying data characteristics, improving the model's adaptability and performance.
(3)Layer-specific and task-specific characteristics of the learnable parameter: We identify that the optimal weights for skip connections vary not only across different layers within a deep network but also across different training tasks. This insight challenges the conventional one-size-fits-all approach of traditional ResNet, where a uniform weight ratio is applied across all layers, regardless of the specific role of each layer or the nature of the training data.

The remainder of this paper is organized as follows. Section II describes the AdaResNet model in detail, including the formulation of the $weight_{tfd}^{ipd}$ parameter and the corresponding backpropagation. Section III presents our experimental setup and results. In Section IV, we review related work in deep learning architectures and adaptive mechanisms.  Finally, Section V concludes the paper.

\section{Model}
In very deep networks, gradients can become extremely small during backpropagation, making it difficult to train the early layers. ResNet (Residual Network) addresses this challenge by allowing gradients to flow directly through the network via skip connections, facilitating the training of much deeper networks.

The process of transforming input data (\(\mathbf{x}\)) to produce output (\(\mathbf{y}\)) in a traditional ResNet can be described as in \eqref{eq:resnet_process}

\begin{align}
    & \mathbf{x} \rightarrow f_1(\mathbf{x}) \rightarrow f_2(f_1(\mathbf{x})) \rightarrow \dots \rightarrow f_n(\dots(f_1(\mathbf{x})\dots)) \notag \\
    & \mathbf{x} \longrightarrow f'(\mathbf{x}) \quad \text{(skip connection)} \notag \\
    & \mathbf{y} = f_{act}(f_n(\dots(f_1(\mathbf{x})\dots)) + f'(\mathbf{x})) \label{eq:resnet_process}
\end{align}

Here, the input \(\mathbf{x}\) is successively transformed by functions \(f_1, f_2, \dots, f_n\). The original input \(\mathbf{x}\) or its less transformed format ($f'(x)$) is then added via a shortcut connection (identity mapping) to the output of the final transformation \(f_n\), producing the final output \(\mathbf{y}\) through the corresponding activation function \(f_{act}\).

In this process, the sequence of transformations \( f_n(\dots(f_1(\mathbf{x})\dots)) \) constitutes the main computational pathway in ResNet, and we refer to its output as the \textbf{transformed data} (\( tfd \)). On the other hand, \( f'(\mathbf{x}) \)or $f'(x)$, which is either less processed or directly the input, is utilized to facilitate the training of much deeper networks, and we refer to it as the \textbf{input represent data} (\( ipd \)) or simply the \textbf{input}.
\\

\begin{definition}
    \emph{Transformed data}. The transformed data refers to the output generated after applying a series of operations—such as convolution, batch normalization, and activation functions—on an input within a residual block of a neural network. This data represents the modifications made to the input as it passes through various layers in the block, capturing the learned features and patterns.
\end{definition}

In a residual block, the transformed data is the result of the main processing path, which typically involves several convolutional layers followed by normalization and activation. This data is then combined with the input represent data (often via addition) to form the output of the residual block, enabling the network to learn more complex functions by effectively adding incremental changes to the input.

\begin{definition}
    \emph{Input represent data}. The input represent data refers to the data that is passed directly from the input of a residual block to its output, often without undergoing significant transformation. This data serves as a baseline or identity mapping, allowing the network to retain and propagate the original input features alongside the transformed features from the main processing path.
\end{definition}

In a residual block, the input represent data typically bypasses the primary convolutional operations and is combined with the transformed data at the block’s output. This bypass, or shortcut connection, helps mitigate issues like vanishing gradients by ensuring that gradients can flow more easily through the network, leading to more effective training of deep models.
\\

The combination of \( f_n(\dots(f_1(\mathbf{x})\dots)) + f'(\mathbf{x}) \) not only facilitates easier propagation of gradients to earlier layers but also impacts the final results differently.

However, the contributions of the input represent data and the transformed data may not be equal.  To control the influence of each component, we introduce a weight between the input and the transformed data, referred to as the \textbf{weight of transformed data and input represent data}. This weight is denoted by the variable \( \textit{weight}_{tfd}^{ipd} \), where \( tfd \) stands for the Transformed Data and \( ipd \) stands for the Input Represent Data.

This approach forms the foundation of the \textbf{AdaResNet} architecture, a variant of the ResNet architecture that incorporates the \( \textit{weight}_{tfd}^{ipd} \) to modulate the contribution of the input. AdaResNet is closely related to ResNet; for example, AdaResNet50 is based on the ResNet50 architecture but includes this weighted mechanism.

In the modified structure, the weight is introduced as shown in Equation \eqref{eq:weight_def}.

\begin{align}
    & \mathbf{x} \rightarrow f_1(\mathbf{x}) \rightarrow f_2(f_1(\mathbf{x})) \rightarrow \dots \rightarrow f_n(\dots(f_1(\mathbf{x})\dots)) \notag \\
    & \mathbf{x} \longrightarrow f'(\mathbf{x}) \quad \text{(skip connection)} \notag \\
    & \mathbf{y} = f_{act}(f_n(\dots(f_1(\mathbf{x})\dots)) + \mathbf{weight}_{tfd}^{ipd} \cdot f'(\mathbf{x})) \label{eq:weight_def}
\end{align}

\begin{figure}
    \includegraphics[width=3.5in]{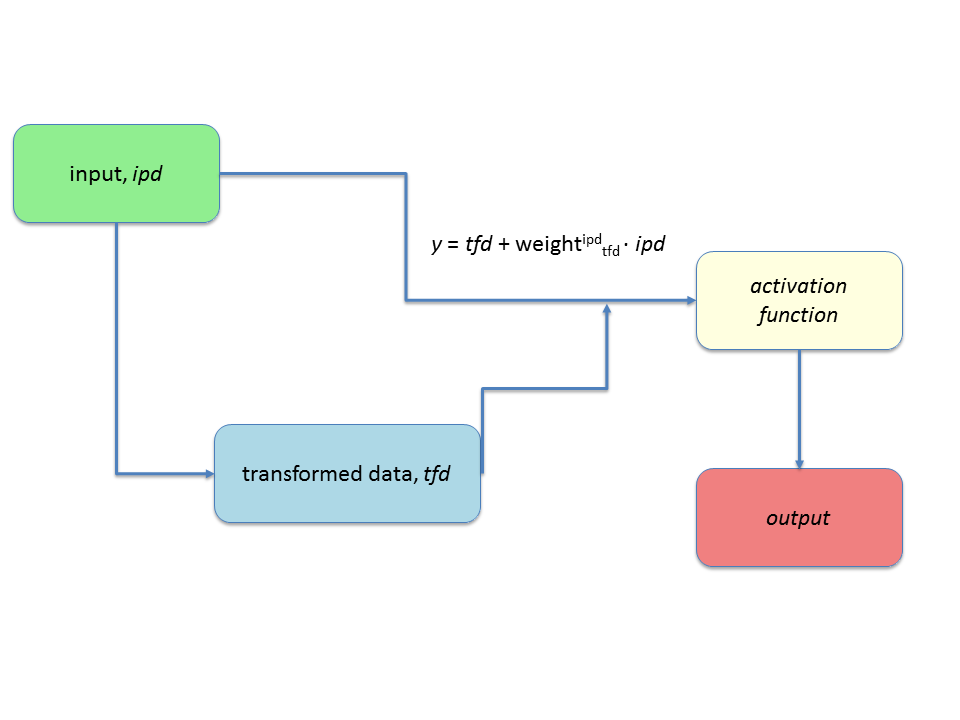}
    \caption{Incorporating weighting into residual learning and blocks}
    \label{beta_diagram}
\end{figure}

The parameter \( \textit{weight}_{tfd}^{ipd} \) enables the network to learn the optimal influence of the input \( \mathbf{x} \) on the final output \( \mathbf{y} \). If \( \textit{weight}_{tfd}^{ipd} \) is learned to be close to zero, the network emphasizes the transformed data \( \mathbf{d} \) over the raw input. Conversely, a larger \( \textit{weight}_{tfd}^{ipd} \) indicates a greater influence of the raw input. When \( \textit{weight}_{tfd}^{ipd} \) equals 1, the model functions as a traditional ResNet.

Additionally, \( \textit{weight}_{tfd}^{ipd} \) is automatically adjusted based on the input data. Specifically, it changes dynamically in response to the loss function during training, being updated through the process of gradient descent. We analyze this process in detail in the following section.

\subsection{Gradient Descent Algorithm}

Assuming the loss function is \( L \), the update formula for the parameter \( \textit{weight}_{tfd}^{ipd} \) during each training step is given by \eqref{weight_change_eq}.

\begin{equation}\label{weight_change_eq}
    \textit{weight}_{tfd}^{ipd} \leftarrow \textit{weight}_{tfd}^{ipd} - \eta \frac{\partial L}{\partial \textit{weight}_{tfd}^{ipd}}
\end{equation}

where:
- \( \eta \) is the learning rate, which controls the step size of the update.
- \( \frac{\partial L}{\partial \textit{weight}_{tfd}^{ipd}} \) is the gradient of the loss function with respect to \( \textit{weight}_{tfd}^{ipd} \).

Using \eqref{weight_change_eq}, \( \textit{weight}_{tfd}^{ipd} \) is gradually adjusted to minimize the loss function. As a result, \( \textit{weight}_{tfd}^{ipd} \) changes dynamically during the training process, enabling the model to better fit the data by optimizing the balance between the input represent data (\( \textit{ipd} \)) and the transformed data (\( \textit{tfd} \)). This automatic adjustment process helps improve the final prediction accuracy.

Below, we provide a detailed description of the backward pass and the updating process for \( \textit{weight}_{tfd}^{ipd} \).

Given the output of the model \( \mathbf{y} = tfd + \textit{weight}_{tfd}^{ipd} \cdot ipd \) (a simplified representation of the ResNet, where a typical ResNet contains multiple instances of this structure), the loss function \( L(\mathbf{y}, \mathbf{y}_{\text{true}}) \) measures the difference between the predicted output \( \mathbf{y} \) and the true labels \( \mathbf{y}_{\text{true}} \).

\subsubsection{Gradient of the Loss Function with Respect to the Output \( \mathbf{y} \)}
During the backward pass, the objective is to compute the gradients of the loss function \( L \) with respect to each of the model's parameters. These gradients are used to update the parameters in the direction that minimizes the loss.

The computation of the gradient (\( gra \)) of the loss function is shown in Equation \eqref{gradient_cal}.

\begin{equation} \label{gradient_cal}
    gra = \frac{\partial L}{\partial \mathbf{y}}
\end{equation}

This gradient indicates how changes in the output \( \mathbf{y} \) of AdaResNet affect the loss function. It is computed by differentiating the loss function with respect to the output \( \mathbf{y} \) of AdaResNet.

\subsubsection{Gradient of the Output \( \mathbf{y} \) with Respect to \( weight_{tfd}^{ipd} \)}

Next, we determine how changes in \( weight_{tfd}^{ipd} \) affect the output \( \mathbf{y} \). Recall that:

\[ \mathbf{y} = tfd + \textit{weight}_{tfd}^{ipd} \cdot ipd \]

Taking the partial derivative of \( \mathbf{y} \) with respect to \( weight_{tfd}^{ipd} \) gives:

\[ \frac{\partial \mathbf{y}}{\partial weight_{tfd}^{ipd}} = ipd \]

Additionally, we can assign a weight to \( tfd \) (the processed intermediary data). However, since this involves a relative relationship between \( tfd \) and \( ipd \), we choose to set \( weight_{tfd}^{ipd} \) relative to \( ipd \).

\subsubsection{Gradient of the Loss Function with Respect to \( weight_{tfd}^{ipd} \)}

By applying the chain rule, the gradient of the loss function \( L \) with respect to \( weight_{tfd}^{ipd} \) is given by:

\[ \frac{\partial L}{\partial weight_{tfd}^{ipd}} = \frac{\partial L}{\partial \mathbf{y}} \cdot \frac{\partial \mathbf{y}}{\partial weight_{tfd}^{ipd}} \]

Substituting the previously computed gradients:

\[ \frac{\partial L}{\partial weight_{tfd}^{ipd}} = \frac{\partial L}{\partial \mathbf{y}} \cdot ipd \]

This gradient demonstrates how changes in \( weight_{tfd}^{ipd} \) will affect the loss function. It is used to update \( weight_{tfd}^{ipd} \) during the optimization step, which will adjust the relative influence between \( tfd \) and \( ipd \).

Although this derivation is based on a simplified form of AdaResNet with a single layer contributing to the output (\( \mathbf{y} = tfd + \textit{weight}_{tfd}^{ipd} \cdot ipd \)), the same principles apply to the full AdaResNet architecture, which may have multiple layers (e.g., \( \mathbf{y} = fu_1(fu_2(\dots fu_n(tfd + \textit{weight}_{tfd}^{ipd} \cdot ipd))) \)).

\subsubsection{Parameter Update}

During the parameter update step, an optimization algorithm (e.g., gradient descent or Adam) uses the computed gradients to update \( weight_{tfd}^{ipd} \). For gradient descent, the update rule is:

\[ weight_{tfd}^{ipd} \leftarrow weight_{tfd}^{ipd} - \eta \frac{\partial L}{\partial weight_{tfd}^{ipd}} \]

where \( \eta \) is the learning rate. This update step is repeated for each batch of training data across multiple epochs, leading to an optimized result from the training data.

\subsection{Training Neural Network with Custom Parameter \( weight_{tfd}^{ipd} \)}
Based on the proposed model and backpropagation mechanism, the training process of AdaResNet is as follows.

\subsubsection{Forward Pass of \( weight_{tfd}^{ipd} \)}
   - During the forward pass, the custom layer receives inputs \( ipd \) and the intermediate result \( tfd \), and then calculates the output as \( tfd + weight_{tfd}^{ipd} \cdot ipd \). This output is then passed to subsequent layers or serves as the final model output.

\subsubsection{Calculating the Loss Function}
   - The model output is compared with the true labels to compute the loss function (assumed to be \( L \)).

\subsubsection{Backward Pass}
   - The backpropagation algorithm calculates the gradients of the loss function with respect to the model parameters. During this process, the gradient of \( weight_{tfd}^{ipd} \) is also computed.

\subsubsection{Updating the Parameters}
   - The optimizer (such as Adam) updates all trainable parameters, including \( weight_{tfd}^{ipd} \), based on the computed gradients. This update process is based on the gradient descent algorithm, causing \( weight_{tfd}^{ipd} \) to adjust slightly after each batch of data to minimize the loss function.

The process of using \( weight_{tfd}^{ipd} \) can be described in Algorithm \ref{alg:train_with_beta}.

\begin{algorithm}
    \caption{Training Neural Network with Custom Parameter \( \textit{weight}_{tfd}^{ipd} \)}
    \label{alg:train_with_beta}
    \begin{algorithmic}[1]
    \REQUIRE Training data \( (X_{\text{train}}, Y_{\text{train}}) \), Testing data \( (X_{\text{test}}, Y_{\text{test}}) \), learning rate \( \eta \), number of epochs \( E \)
    \ENSURE Trained model with optimized \( \textit{weight}_{tfd}^{ipd} \)
    \STATE Initialize model parameters, including \( \textit{weight}_{tfd}^{ipd} \)
    \STATE Normalize \( X_{\text{train}} \) and \( X_{\text{test}} \)
    \STATE Convert \( Y_{\text{train}} \) and \( Y_{\text{test}} \) to one-hot encoding

    \FOR{epoch = 1 to \( E \)}
        \FOR{each batch \( (X_{\text{batch}}, Y_{\text{batch}}) \) in \( (X_{\text{train}}, Y_{\text{train}}) \)}
            \STATE \( $tfd$ \gets \text{ResNet50\_intermediate}(X_{\text{batch}}) \)
            \STATE \( O \gets \text{AddWithWeight}($tfd$, X_{\text{batch}}, \textit{weight}_{tfd}^{ipd}) \)
            \STATE predictions \( \gets \text{Model}(O) \)
            \STATE \( L \gets \text{ComputeLoss}(\text{predictions}, Y_{\text{batch}}) \)
            \STATE gradients \( \gets \text{Backpropagate}(L, \text{parameters}) \)
            \STATE \text{UpdateParameters}(\text{parameters}, \text{gradients}, $\eta$)
        \ENDFOR
    \ENDFOR

    \STATE Evaluate the model on \( X_{\text{test}} \) and \( Y_{\text{test}} \)
    \STATE \RETURN trained model
    \end{algorithmic}
\end{algorithm}

\subsection{Brief Explanation}

In this subsection, we briefly explain the rationale for introducing the weight between the Transformed Data and the Input Represent Data.

In the equation \( f_n(\dots(f_1(\mathbf{x})\dots)) + f'(\mathbf{x}) \), \( f'(\mathbf{x}) \) inherently contributes equally to the output as \( f_n(\dots(f_1(\mathbf{x})\dots)) \), meaning that both have the same impact on the final output. However, in most cases, we cannot assume this equal contribution. Even within the same scenario, different training data can alter the relationship between these contributions.

To formalize this, we introduce a function \( \textit{contrib}(p, r) \) to describe how much a parameter \( p \) contributes to the output \( r \). In ResNet, both the input represent data \( \textit{ipd} \) and the transformed data \( \textit{tfd} \) contribute to the recognition target \( r \). However, in general, we cannot assume that \( \textit{contrib}(\textit{ipd}, r) = \textit{contrib}(\textit{tfd}, r) \).

We use a counterexample to illustrate the need for variable weighting. Assume that the input data \( f'(\mathbf{x}) \) has the same weight as the intermediate results. One key feature of ResNet is that it can be extended into many layers. Let us consider two consecutive layers, \( n \) and \( n+1 \), and examine the contributions \( \text{contrib}(f_n(\dots(f_1(\mathbf{x}))\dots)) \) and \( \text{contrib}(f_{n+1}(\dots(f_1(\mathbf{x}))\dots)) \).

If \( \text{contrib}(f_n(\dots(f_1(\mathbf{x}))\dots)) = \text{contrib}(x_{n}) \) in layer $n$, where \( x_{n} \) represents the input of the \( n \) layer, then when the process continues to the next layer \( n+1 \), the input data is now \( x_{n+1} \), and the transformed data is \( f_{n+1}(\dots(f_1(\mathbf{x}))\dots) \). The input data of layer \( n+1 \) is derived from the processed results of layer \( n \), and since \( x_{n} \) has undergone non-linear processing (e.g., through the ReLU activation function) in layer $n$, it is difficult to maintain a linear one-to-one relationship between the input data and the transformed data. Therefore, there is no guarantee that the contributions will remain equal in layer $n+1$, as shown in \eqref{assu_equ_n}. In fact, as the number of layers increases, it becomes more likely that their contributions will diverge.

\begin{align}\label{assu_equ_n}
    &\textit{if \ \ \ } \text{contrib}(f_n(\dots(f_1(\mathbf{x}))\dots)) = \text{contrib}(x_n), \notag \\
    &\textit{then } \text{contrib}(f_{n+1}(\dots(f_1(\mathbf{x}))\dots)) \neq \text{contrib}(x_{n+1}).
\end{align}

We conclude, as shown in \eqref{assu_equ_n}, that in most cases, even if one layer exhibits equal contributions from the input and the transformed data, it is unlikely that all layers will maintain this equality. Consequently, the weights cannot be assumed to be equal across the network.

Therefore, \( \textit{weight}_{tfd}^{ipd} \) must be adjusted during the learning process, meaning it should \textbf{dynamically change} throughout training. This dynamic adjustment is crucial for ensuring that the network can effectively capture and utilize relevant features while minimizing the impact of irrelevant or noisy data.

\subsection{Factors influencing \( \textit{weight}_{tfd}^{ipd} \)}
Several factors influence \( \textit{weight}_{tfd}^{ipd} \), including:

\subsubsection{Dependency on Training Datasets}
The first challenge is that \( \textit{weight}_{tfd}^{ipd} \) can vary significantly depending on the specific training dataset used. Different datasets possess unique distributions and characteristics, necessitating the adaptation of \( \textit{weight}_{tfd}^{ipd} \) to ensure optimal performance. 

\begin{equation}
    \frac{\text{contrib}(f_n(\dots(f_1(\mathbf{x}))\dots))}{\text{contrib}(f_n(\mathbf{x}))} \text{ is related to } \text{data}_{\text{train}}
\end{equation}
where \( \text{data}_{\text{train}} \) represents sub sets of a training dataset.
\\

Moreover, this ratio often differs when training on different datasets, such as MNIST and CIFAR-10.
\begin{equation}
    \frac{\text{contrib}(f_n(\dots(f_1(\mathbf{x}))\dots))}{\text{contrib}(f_n(\mathbf{x}))} \text{ is related to } \text{type}_{\text{train}}
\end{equation}
where \( \text{data}_{\text{train}} \) represents type of the training datasets.
\\

\subsubsection{Neural Network Architecture}
The specific neural network architecture also plays a significant role in determining the optimal value of \( \textit{weight}_{tfd}^{ipd} \). Networks with varying depths, widths, and connectivity patterns exhibit distinct learning behaviors, thereby affecting the sensitivity and responsiveness of \( \textit{weight}_{tfd}^{ipd} \) to changes in the training data. Consequently, the dynamic adjustment of \( \textit{weight}_{tfd}^{ipd} \) must be tailored to the specific architecture of the neural network in question.

In a ResNet network, there are different stages (such as several identity blocks and convolutional blocks\footnote{\url{https://github.com/keras-team/keras-applications/blob/master/keras\_applications/resnet50.py}}, each of which can be seen as a stage) to use the \(weight_{tfd}^{ipd}\), those values can also be different in different stage. Thus, to reflect difference in each stage, the weight can be an array with respect to each stage as in \eqref{weight_as_array}. This may increase the complex of neural network, for simplicity, \(weight_{tfd}^{ipd}\) can be a unique one in all stages in some scenarios.

\begin{equation} \label{weight_as_array}
    weight_{tfd}^{ipd} = \{weight_{tfd}^{ipd\_1}, weight_{tfd}^{ipd\_2}, ..., weight_{tfd}^{ipd\_i}, ... \}
\end{equation}
,where $i$ are stage of a neural network. Stage is one place where make input data to be mixed with processed data, such as one identify block or one cov block.

\subsubsection{Non-Uniqueness of Optimal Values}
A further challenge lies in the fact that the optimal value of \(weight_{tfd}^{ipd}\) may not be unique, but rather exist as a set of potential values. This non-uniqueness stems from the inherent complexity and redundancy within neural networks, which often possess multiple solutions that achieve similar levels of performance. 

\section{Verification}
To validate the effectiveness of the proposed method, we conducted comparative experiments using three different approaches: (1) the proposed method based on ResNet 50 with a trainable weight, AdaResNet (2) the traditional ResNet 50, and (3) a method using a fixed weight (2x) instead of a trainable one. The results over 10 epochs are reported and discussed.

\subsection{Accuracy}
\subsubsection{Experimental Setup}
The model was trained and evaluated on the CIFAR-10 dataset with ResNet50, as the accuracy by this model is about 40\%, which can have enough space to show whether there are some improvement or not (On the other hand, if we use MNIST dataset, its accuracy can achieve more than 99\%, if there are some improvement, it still small). The dataset consists of 60,000 32x32 color images in 10 classes, with 50,000 training images and 10,000 test images. The images were normalized to a range of [0, 1] and the labels were converted to one-hot encoded vectors.

In this verification, ResNet and AdaResNet are compared. For ResNet, we use the Keras library of TensorFlow. AdaResNet are customed based on the Keras library of TensorFlow too.
It is a custom ResNet model modified to incorporate a trainable parameter \( weight_{tfd}^{ipd} \) that scales the input before adding it to an intermediate feature map. This modification is intended to examine the impact of dynamic feature scaling on the performance of the model when applied to the CIFAR-10 dataset. The model is constructed using the Keras framework, and the details of the implementation are outlined below.

The implementation includes creating a custom layer within the Keras framework that integrates the trainable parameter \( \textit{weight}_{tfd}^{ipd} \). The setup process is summarized in Algorithm 1. For more detailed information, please refer to the code available on GitHub\footnote{https://github.com/suguest/AdaResNet}.

\begin{algorithm}[H]
\caption{Integration of Trainable Parameter \( weight_{tfd}^{ipd} \) in ResNet}
\begin{algorithmic}[1]
\REQUIRE CIFAR-10 dataset \( \mathcal{D} \) with training set \( \mathcal{D}_{\text{train}} \) and test set \( \mathcal{D}_{\text{test}} \)
\ENSURE Trained ResNet model with dynamic parameter \( weight_{tfd}^{ipd} \)

\STATE \textbf{Step 1: Load and Preprocess Data}
\STATE Normalize images in \( \mathcal{D}_{\text{train}} \) and \( \mathcal{D}_{\text{test}} \) to [0, 1]
\STATE Convert labels in \( \mathcal{D}_{\text{train}} \) and \( \mathcal{D}_{\text{test}} \) to one-hot encoding

\STATE \textbf{Step 2: Define Custom Layer with Trainable \( weight_{tfd}^{ipd} \)}
\STATE Initialize \( weight_{tfd}^{ipd} \) as a trainable parameter with an initial value of 0
\STATE Define forward pass as \( \mathbf{y} = \mathbf{d} + weight_{tfd}^{ipd} \cdot \mathbf{x} \)

\STATE \textbf{Step 3: Construct ResNet Model}
\STATE Import ResNet50 architecture without the top classification layer
\STATE Extract intermediate feature map \( \mathbf{d} \)
\STATE Apply custom layer to compute \( \mathbf{y} \)

\STATE \textbf{Step 4: Compile Model}
\STATE Use Adam optimizer with categorical cross-entropy loss
\STATE Define accuracy as the evaluation metric

\STATE \textbf{Step 5: Train and Evaluate Model}
\STATE Train model on \( \mathcal{D}_{\text{train}} \) for 10 epochs
\STATE Evaluate model on \( \mathcal{D}_{\text{test}} \)
\STATE Output model accuracy and loss
\end{algorithmic}
\end{algorithm}

The experiments were performed on the CIFAR-10 dataset. Each method was trained for 10 epochs, and the performance metrics such as accuracy and loss were recorded for both training and validation datasets. Below are the verification results.

\subsubsection{Results}
\begin{figure}
    \includegraphics[width=3.5 in]{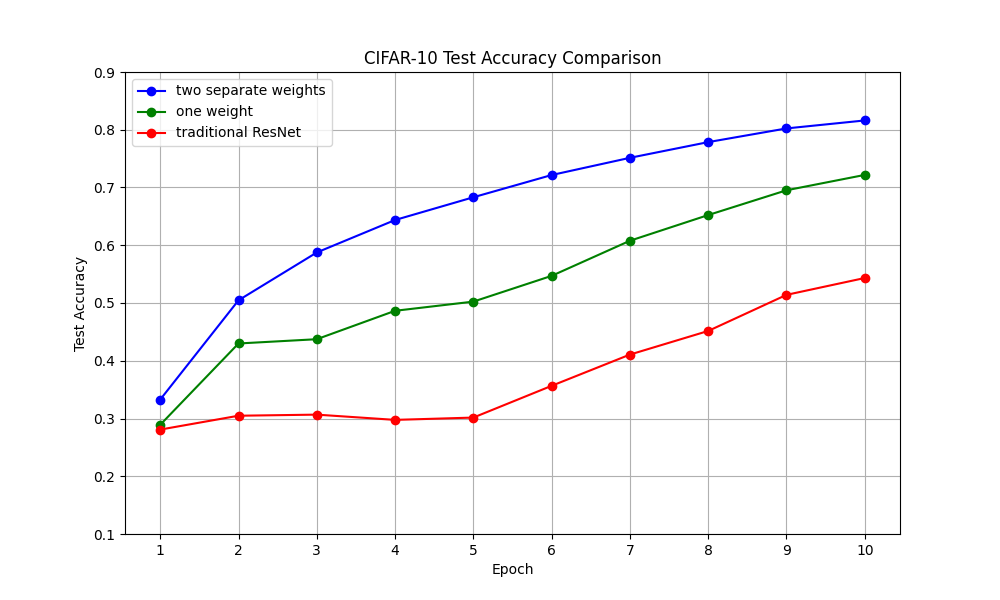}
    \caption{Comparison of training accuracy}
    \label{accuracy_comparison}
\end{figure}

\begin{figure}
    \includegraphics[width = 3.6 in]{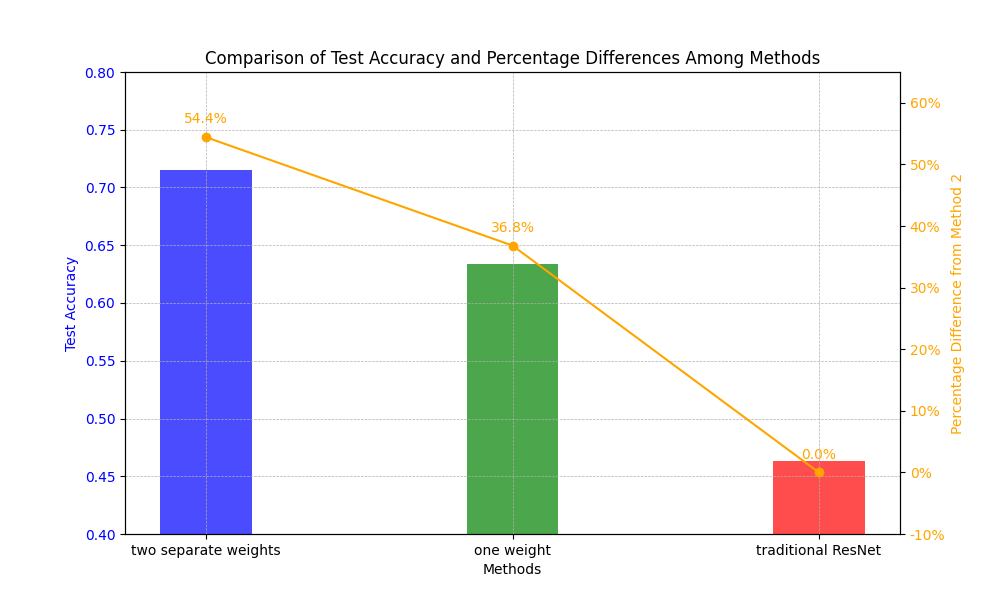}
    \caption{Comparison of test accuracy}
    \label{testAccuracyAndDiff}
\end{figure}

The professional learning curves is shown in Figure \ref{accuracy_comparison} and Figure \ref{testAccuracyAndDiff} that illustrate the training and validation accuracy for each method over the 10 epochs.

The comparison clearly demonstrates the differences among the three methods.
The two methods of AdaResNet show higher accuracy in both accuracies on the training data and test data.
For the training data, AdaResNet achieves the highest final test accuracy of 0.81 and 0.72 separately for AdaResNet with two weights and one unified weight, which has more than 0.26 and 0.18 increase in the accuracy than the traditional ResNet method with a accuracy of  0.46.
For the test data, the proposed method show an accuracy of 0.71 and 0.63 for two methods of AdaResNet, which has a more accuracy of than 0.25 and 0.18 than that of the traditional method (0.46). The AdaResNet with two separate weights has an increase of \textbf{54.35\% increase} of traditional ResNet.

When comparison of two methods of AdaResNet, one with the unified weight and another with separate weights, the method with separate weights has more accuracy improvement. This indicates that there are different relationship among the input and intermediate process results between the identify block and conv block.

From the above results, it indicates that the trainable weight effectively balances the influence of the raw input and the transformed data, leading to improved learning and generalization.

\subsection{Weights Impact}
In this section, we aim to verify that $weight_{tfd}^{ipd}$ is a dynamic parameter rather than a fixed value, capable of adapting to different training tasks and varying even within the same task across different training iterations.

For a better comparison, we output the $weight_{tfd}^{ipd}$ after each training is done, i.e. to iterate to output the  $weight_{tfd}^{ipd}$ of each layer, as shown in \ref{output_weight_al}.

\begin{algorithm}
    \caption{Algorithm to output the final $weight_{tfd}^{ipd}$}
    \label{output_weight_al}
    \begin{algorithmic}[1]
        \FOR{each layer in the model}
            \IF{layer is of type \texttt{customLayer}}
                \STATE Retrieve the weight value from the layer
                \STATE Output the layer's name and the corresponding weight value
            \ENDIF
        \ENDFOR
        \end{algorithmic}
\end{algorithm}

\subsubsection{Whether $weight_{tfd}^{ipd}$ is a single value or not}
\begin{figure*}[htb]
    \centering
    \includegraphics[width=6.5in]{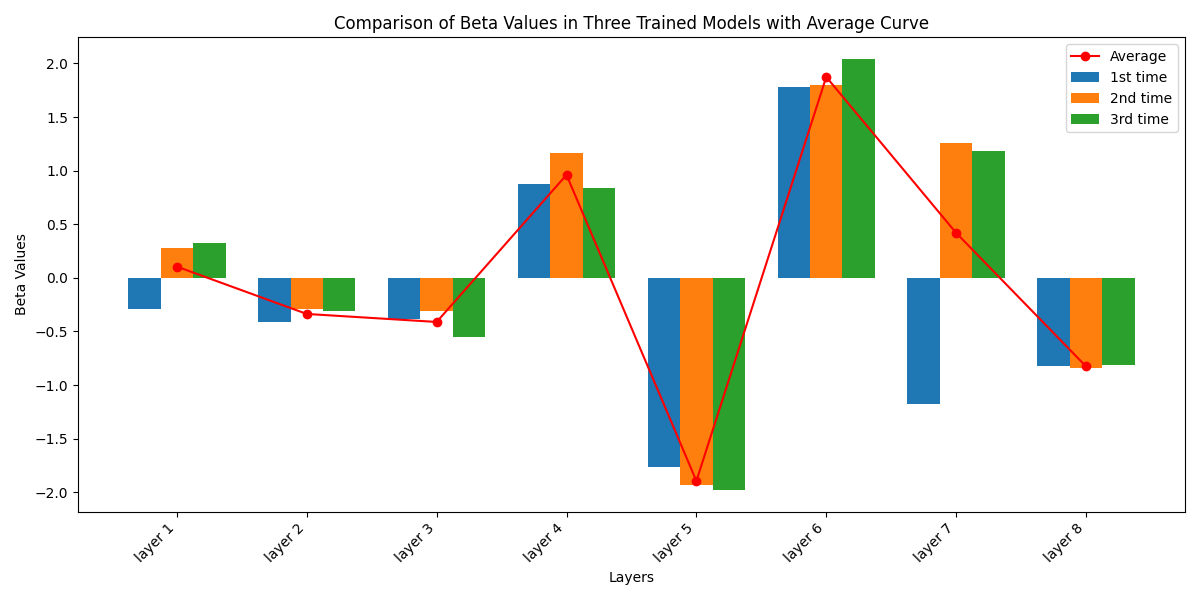}
    \caption{Weights of different layers}
    \label{weightComparison}
\end{figure*}

In this subsection, we aim to determine whether the $weight_{tfd}^{ipd}$ remains consistent across runs. We conducted three separate runs of AdaResNet, all starting with the same initial parameters and using the same training data (CIFAR-10), with each model trained for 10 epochs. The results are shown in Figure \ref{weightComparison} and table \ref{tab:weights}.

\begin{table}[htb]
    \centering
    \caption{Weights in Different Layers for Three Rounds of Testing (cifar10)}
    \begin{tabular}{|c|c|c|c|}
    \hline
    \textbf{Layer} & \textbf{round\_1} & \textbf{round\_2} & \textbf{round\_3} \\ \hline
    1 & -0.28722298 & 0.27989703 & 0.32219923 \\ \hline
    2 & -0.41371468 & -0.28776032 & -0.30848 \\ \hline
    3 & -0.37947246 & -0.3051696 & -0.5491747 \\ \hline
    4 & 0.8734257 & 1.1673123 & 0.84171796 \\ \hline
    5 & -1.7672663 & -1.9361044 & -1.9803141 \\ \hline
    6 & 1.7821076 & 1.7983766 & 2.0427594 \\ \hline
    7 & -1.1800854 & 1.2597568 & 1.1798627 \\ \hline
    8 & -0.82326496 & -0.8402289 & -0.8131428 \\ \hline
    \end{tabular}
    \label{tab:weights}
\end{table}

\begin{figure}
    \centering
    \includegraphics[width=3.5in]{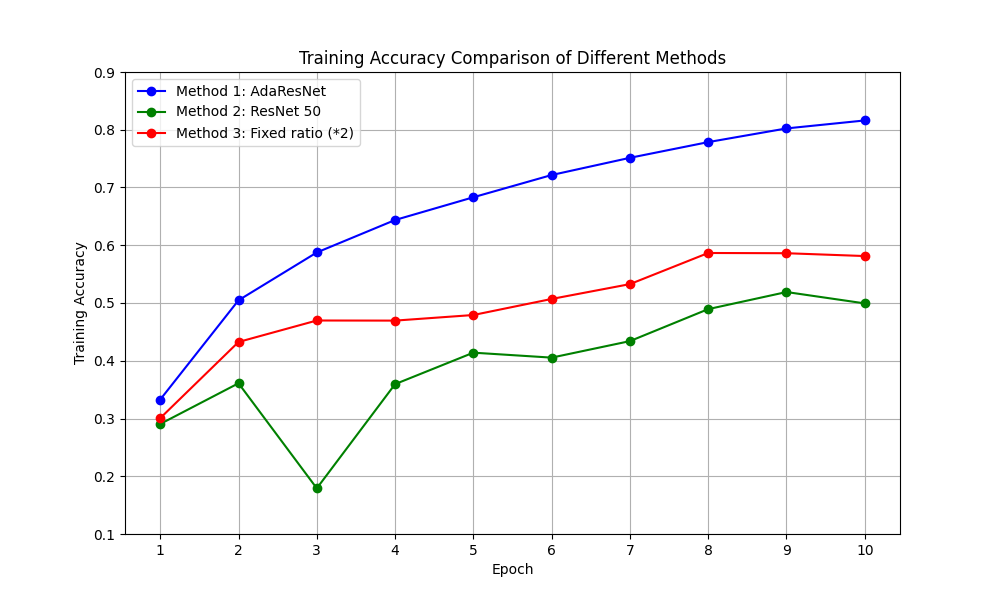}
    \caption{Accuracy comparison of fixed weight}
    \label{comparisonOfFixedWeight}
\end{figure}

From Figure \ref{weightComparison}, we can see that the weights values are different in different layers. This indicates that it is not suitable to use a fixed value for the combination of input and the intermediately processed data. We also combined to use a fixed ratio among the input data the intermediately processed data of 2 as shown in Figure \ref{comparisonOfFixedWeight}, which also shows a higher accuracy than to use the dynamic  $weight_{tfd}^{ipd}$.

The difference of weight in different test rounds can also be seen in Table \ref{tab:weights}. The weight values across the three test rounds exhibit variations, indicating that the weights differ between layers. For instance, at layer 1, the weights are -0.2872, 0.2799, and 0.3222 for rounds one, two, and three, respectively, demonstrating a significant range of approximately 0.61 between the lowest and highest values. Similarly, at layer 5, the weights are -1.7673, -1.9361, and -1.9803, again showing variability with a difference of about 0.21. These differences underscore that the weights in each layer are not consistent across different rounds of testing, which can be attributed to factors such as random initialization and the stochastic nature of the training process.

\subsubsection{Whether $weight_{tfd}^{ipd}$ is different among different training task}
\begin{figure*}[htb]
    \centering
    \includegraphics[width=6.5in]{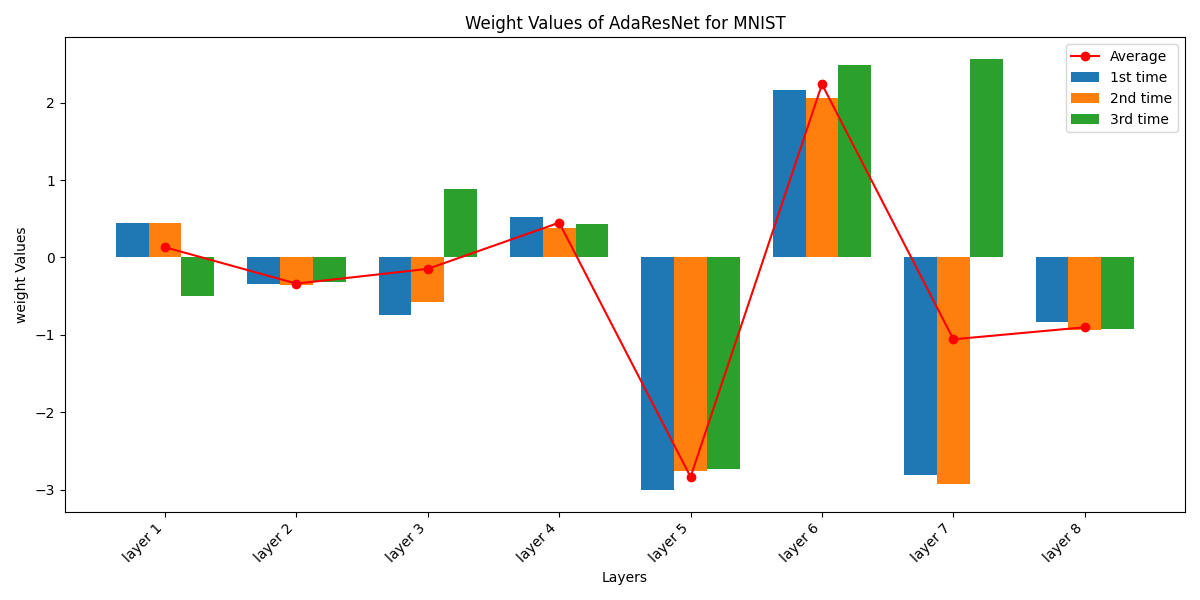}
    \caption{Weights of different layers for MNIST}
    \label{weightComparisonMNIST}
\end{figure*}

While for different classification tasks, such as for MNIST, the weights have big difference. For the weights of MNIST, we also carry out three verification, the results are shown in Figure \ref{weightComparisonMNIST} and table \ref{tab:mnist_weight_cmp}.

\begin{table}[htb]
    \centering
    \caption{Weights in Different Layers for Three Rounds of Testing (MNIST)}
    \begin{tabular}{|c|c|c|c|}
    \hline
    \textbf{Layer} & \textbf{round\_1} & \textbf{round\_2} & \textbf{round\_3} \\ \hline
    1 & 0.44887054 & 0.4484792 & -0.5003674 \\ \hline
    2 & -0.34602356 & -0.35169616 & -0.31584582 \\ \hline
    3 & -0.74334604 & -0.5807008 & 0.8818225 \\ \hline
    4 & 0.5266892 & 0.3835334 & 0.43830293 \\ \hline
    5 & -3.0067017 & -2.7609563 & -2.7376952 \\ \hline
    6 & 2.1653237 & 2.065729 & 2.4824123 \\ \hline
    7 & -2.8167214 & -2.9216428 & 2.5657778 \\ \hline
    8 & -0.8365008 & -0.94025135 & -0.9289533 \\ \hline
    \end{tabular}
    \label{tab:mnist_weight_cmp}
\end{table}

Thus, we analyze the difference between within group and between groups. The variance of two groups of weight data, representing the CIFAR-10 and MNIST datasets, was analyzed using absolute values. The CIFAR-10 and MNIST groups each comprised three sets of eight weights. The within-group variance was computed by averaging the variance across the corresponding columns within each group, while the between-group variance was calculated by assessing the variance between the mean values of the columns across the two groups.

The results revealed a within-group variance of 0.0113 for the CIFAR-10 group and 0.0074 for the MNIST group, indicating that the CIFAR-10 group exhibits slightly higher variability among its data points compared to the MNIST group. Furthermore, the between-group variance was calculated to be 0.1205, which is significantly higher than both within-group variances. This suggests that the differences between the mean values of the CIFAR-10 and MNIST groups are more pronounced than the variations observed within each group. Overall, the analysis highlights that the between-group differences are more substantial than the differences within the individual groups, with CIFAR-10 showing a marginally greater degree of internal variability than MNIST.

\section{Related Work}
The development of deep neural networks has been one of the most significant advancements in artificial intelligence, with ResNet (Residual Network) standing out as a groundbreaking architecture. Since its introduction by He et al. in 2016 \cite{he2016deep}, ResNet has become a cornerstone in the design of deep networks, particularly for tasks in computer vision such as image classification, object detection, and segmentation.

\subsection{Residual Networks and Skip Connections}
The concept of residual learning was introduced to address the degradation problem in deep neural networks, where adding more layers to a network does not necessarily lead to better performance and often results in higher training error. ResNet's innovative use of skip connections allows the network to learn residual mappings instead of directly learning unreferenced functions \cite{amelio2023representation}. This approach effectively mitigates the vanishing gradient problem, as gradients can propagate more easily through the network. The original ResNet paper demonstrated that networks with over 100 layers could be trained successfully \cite{borawar2023resnet}, a feat previously unattainable with traditional deep architectures.

While ResNet has achieved remarkable success, several extensions and modifications have been proposed to further enhance its performance. For example, Wide ResNet \cite{xu2023resnet} \cite{peng2023ensemble} explores the effect of increasing the width of the network (i.e., the number of channels) instead of just depth, leading to improved performance on various datasets. Another variation, ResNeXt \cite{xie2017aggregated}, introduces a cardinality dimension, allowing for a more flexible combination of feature maps, which has been shown to improve accuracy and efficiency.

\subsection{Adaptive Mechanisms in Neural Networks}
The idea of incorporating adaptive mechanisms into neural networks has gained traction as researchers seek to make models more flexible and responsive to varying data distributions. Squeeze-and-Excitation Networks (SENet) \cite{jin2022delving}, for instance, adaptively recalibrate channel-wise feature responses by explicitly modeling interdependencies between channels. This enables the network to focus on the most informative features, leading to significant performance gains in image recognition tasks.

Another line of research focuses on adaptive learning rates and weights within networks. For example, the use of adaptive learning rates in algorithms such as Adam \cite{chandriah2021rnn} and RMSprop \cite{shi2021rmsprop} has become standard practice in training deep networks, allowing for faster convergence and better generalization.

However, adaptive mechanisms within the architecture itself, such as the one proposed in our AdaResNet, are less explored. Existing methods typically focus on global adjustments, such as learning rates, rather than on dynamically altering the flow of information within the network. The Dynamic Convolution \cite{chen2020dynamic} approach is a notable exception, where convolutional kernels are dynamically adjusted based on input features. However, it does not address the specific challenges posed by skip connections in residual networks.

\subsection{ Limitations of Traditional Residual Networks}
Despite the successes of ResNet and its variants, the uniform treatment of the input ($ipd$) and processed data ($tfd$) in skip connections remains a limitation. Traditional ResNet adds $ipd$ and $tfd$ without considering the varying importance of these components across different layers or training data conditions. This uniformity can lead to suboptimal performance, especially in cases where the relative importance of $ipd$ and $tfd$ differs significantly.

To address this issue, several approaches have been proposed to modify the skip connections in ResNet. For example, the Mixed-Scale Dense Network (MSDNet) \cite{veit2016residual} adapts the receptive field sizes across the network but does not dynamically adjust the skip connections themselves. Similarly, Highway Networks \cite{zilly2017recurrent} introduce gates to control the flow of information through the network, but these gates are static once trained and do not adapt during training.

\subsection{Our Contribution}
Our proposed AdaResNet builds on this body of work by introducing an adaptive mechanism specifically for skip connections in residual networks. By allowing the ratio of $ipd$ to $tfd$, represented by the learnable parameter $weight_{tfd}^{ipd}$, to be adjusted dynamically during training, AdaResNet provides a more flexible and data-responsive architecture. This approach not only addresses the limitations of traditional ResNet but also leverages the strengths of adaptive learning to enhance performance across a range of tasks and datasets.

In summary, while significant progress has been made in the design and optimization of deep neural networks, the uniform treatment of skip connections in residual networks presents a limitation that has yet to be fully addressed. AdaResNet represents a novel contribution in this area, introducing a dynamic and adaptive approach to residual learning that we believe will offer significant benefits in terms of both accuracy and generalization.

\section{Conclusion}
In this paper, we introduced AdaResNet, a novel extension of the ResNet architecture that incorporates an adaptive mechanism for dynamically balancing the contributions of skipped input ($ipd$) and processed data ($tfd$). Traditional ResNet models rely on a fixed 1:1 ratio for combining $ipd$ and $tfd$, which can be suboptimal in various training scenarios. AdaResNet addresses this limitation by introducing a learnable parameter, \(weight_{tfd}^{ipd}\), which is automatically optimized during training. This allows the network to adjust the ratio between $ipd$ and $tfd$ in response to the specific characteristics of the data, thereby enhancing the model's adaptability and overall performance.

Our experimental results demonstrate that AdaResNet consistently outperforms the traditional ResNet architecture, particularly in tasks where the relative importance of $ipd$ and $tfd$ varies across different layers and datasets. We also highlighted the critical insight that the optimal weights for skip connections differ across layers and tasks, challenging the conventional approach of using a uniform weight ratio across the entire network.
By leveraging adaptive skip connections, AdaResNet not only improves accuracy and efficiency but also offers a more nuanced and flexible approach to deep network design. This work opens up new possibilities for further exploration of adaptive mechanisms in neural networks, with potential applications across various domains in deep learning.

Future work will focus on extending the AdaResNet framework to other network architectures and exploring the impact of adaptive mechanisms in different types of neural networks, such as those used in natural language processing and reinforcement learning. Additionally, we plan to investigate the theoretical underpinnings of adaptive skip connections to better understand their role in improving network generalization and robustness.

\section*{Acknowledgment}
The authors thank the anonymous reviewers for their constructive comments, which help us to improve the quality of this paper. This work was supported in part by the National Natural Science Foundation of China under Grant No. 61772352; the Science and Technology Planning Project of Sichuan Province under Grant No. 2019YFG0400, 2018GZDZX0031, 2018GZDZX0004, 2017GZDZX0003, 2018JY0182, 19ZDYF1286.

\ifCLASSOPTIONcaptionsoff
  \newpage
\fi



%

\bibliographystyle{IEEEtran}
\bibliography{ref}

%




\begin{IEEEbiography}{Hong Su}
    Hong Su received the BS and MS degrees, in 2003 and 2006, respectively, from Sichuan University, Chengdu, China. He is currently a researcher of Chengdu University of Information Technology, Chengdu, China. His research interests include blockchain and the Internet of Value.
\end{IEEEbiography}



\end{document}